\documentclass[final]{cvpr}

\usepackage{times}
\usepackage{epsfig}
\usepackage{graphicx}
\usepackage{amsmath}
\usepackage{amssymb}
\usepackage{graphicx}
\usepackage{amsmath}
\usepackage{subcaption}
\usepackage{amsfonts}
\usepackage{adjustbox}
\usepackage{multirow}
\usepackage{pifont}
\usepackage{enumitem}
\newcommand{\PQ}{PQ}
\newcommand{\PQda}{PQ\textsuperscript{$\dagger$}}
\newcommand{\RQ}{RQ}
\newcommand{\SQ}{SQ}
\newcommand{\PQth}{PQ\textsuperscript{Th}}
\newcommand{\RQth}{RQ\textsuperscript{Th}}
\newcommand{\SQth}{SQ\textsuperscript{Th}}
\newcommand{\PQst}{PQ\textsuperscript{St}}
\newcommand{\RQst}{RQ\textsuperscript{St}}
\newcommand{\SQst}{SQ\textsuperscript{St}}
\newcommand{\miou}{mIoU}
\usepackage[usenames,dvipsnames]{xcolor}
\definecolor{dgreen}{rgb}{0.0,0.6,0.0}
\newcommand{\cmark}{\textcolor{dgreen}{\text{\ding{51}}}}%

\usepackage[pagebackref=true,breaklinks=true,colorlinks,bookmarks=false]{hyperref}

\def\cvprPaperID{2936} % *** Enter the CVPR Paper ID here
\def\confYear{CVPR 2021}

\begin{document}

%%%%%%%%% TITLE
\title{Cylindrical and Asymmetrical 3D Convolution Networks\\ for LiDAR Segmentation}

\author{Xinge Zhu$^{\dag*}$~~~ Hui Zhou$^{\dag*}$~~~ Tai Wang$^{\dag}$~~~ Fangzhou Hong$^{\ddagger}$\\ Yuexin Ma$^{\S}$~~~ Wei Li$^{\wr}$~~~ Hongsheng Li$^{\dag}$~~~ Dahua Lin$^{\dag}$ \and {\small $^{\dag}$Chinese University of Hong Kong~~~$^{\S}$ShanghaiTech University} \\ {\small$^{\wr}$Inceptio~~~$^{\ddagger}$Nanyang Technological University}}

\maketitle

\begin{abstract}
    
State-of-the-art methods for large-scale driving-scene LiDAR  segmentation often project the point clouds to 2D space and then process them via 2D convolution. 
Although this corporation shows the competitiveness in the point cloud, it inevitably alters and abandons the 3D topology and geometric relations. A natural remedy is to utilize the 3D voxelization and 3D convolution network.
However, we found that in the outdoor point cloud, the improvement obtained in this way is quite limited. An important reason is the property of the outdoor point cloud, namely sparsity and varying density.
% A straightforward {solution} to tackle the issue of 3D-to-2D projection is to keep the 3D representation and process the points in the 3D space. 
% In this work, we first perform an in-depth analysis for different representations and backbones in 2D and 3D spaces, and reveal the effectiveness of 3D representations and networks on LiDAR segmentation. Specifically, w
Motivated by this investigation, we propose a new framework for the outdoor LiDAR segmentation, where cylindrical partition and asymmetrical 3D convolution networks are designed to explore the 3D geometric pattern while maintaining these inherent properties. 
Moreover, a point-wise refinement module is introduced to alleviate the interference of lossy voxel-based label encoding.
We evaluate the proposed model on two large-scale datasets, \ie, SemanticKITTI and nuScenes. Our method achieves the \textbf{1st place} in the leaderboard of SemanticKITTI~\footnote{The method is in the 1st place before CVPR deadline. $^*$ denotes the equal contribution and code at \url{https://github.com/xinge008/Cylinder3D}. Corresponding email: zhuxinge123@gmail.com} and outperforms existing methods on nuScenes with a noticeable margin, about 4\%. Furthermore, the proposed 3D framework also generalizes well to LiDAR panoptic segmentation and LiDAR 3D detection.
% which exploits the 3D topology relations and structures of driving-scene point clouds. 
% Moreover, a dimension-decomposition based context modeling module is introduced to explore the high-rank context information in point clouds in a progressive manner.
% We evaluate the proposed model on a large-scale driving-scene dataset, \ie~SemanticKITTI. Our method achieves state-of-the-art performance and outperforms existing methods by 6\% in terms of mIoU.
    
\end{abstract}

%===============================================================================

\section{Introduction}
	 3D {LiDAR} sensor has become an {indispensable device} in {modern} autonomous driving {vehicles}. It {captures} more precise and farther-away {distance measurements of the surrounding environments} than conventional visual cameras. The measurements of the sensor naturally form 3D point clouds that can be used to realize a thorough scene understanding for autonomous driving planning and execution,
	 in which LiDAR segmentation is crucial for driving-scene understanding. It aims to identify the pre-defined categories of each 3D point, such as {car, truck, pedestrian,} \etc, which provides point-wise perception information of the overall 3D scene.
	
	 %Prior point cloud based algorithms for autonomous driving can be divided into two major categories: Lidar Detection \cite{yan2018second, shi2019pointrcnn, lang2019pointpillars} and LiDAR Segmentation \cite{wu2018squeezeseg, hu2020randla, milioto2019rangenet++, zhang2020polarnet, wong2020identifying, cortinhal2020salsanext}. Lidar detection only assume each object can be represented by bounding box easily. In \cite{geiger2012we}, 3D object detection benchmark includes three classes: car, pedestrian and cyclist. However, there exist a large amount of stuff in the nature, such as drivable road, parking areas and sidewalks\cite{behley2019semantickitti}. Lidar segmentation can detect things and stuff simultaneously and provide fine-grained scene understanding for decision-control module in autonomous driving. Thus, LiDAR segmentation are quite popular in autonomous driving.
    
    \begin{figure*}
    \centering
    \begin{subfigure}[t]{0.6\textwidth}
    \centering
    \includegraphics[width=1\linewidth]{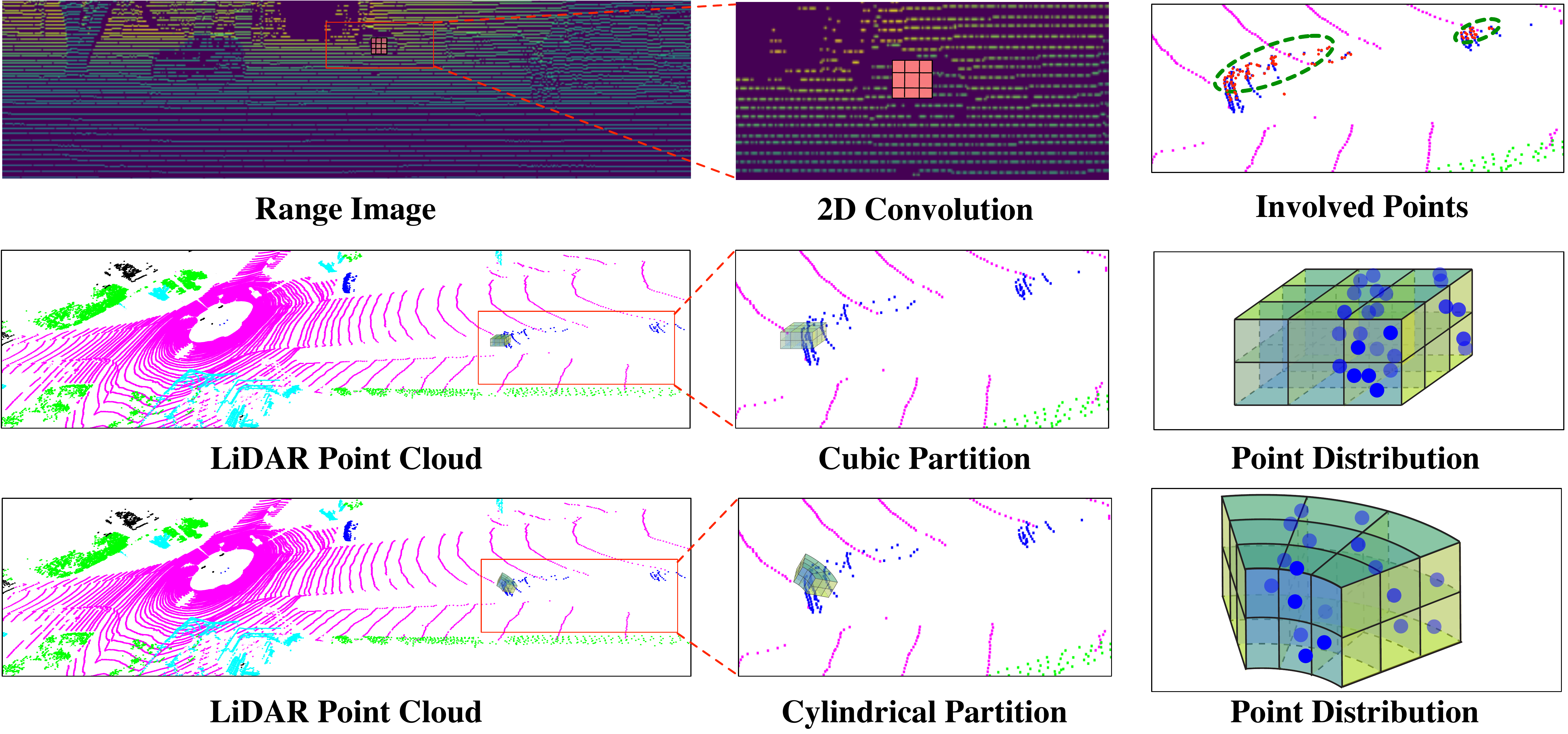}
    \caption{}
    \label{fig:2d_3d_conv}
    \end{subfigure}
    \hfil
    \begin{subfigure}[t]{0.3\textwidth}
        \includegraphics[width=\linewidth]{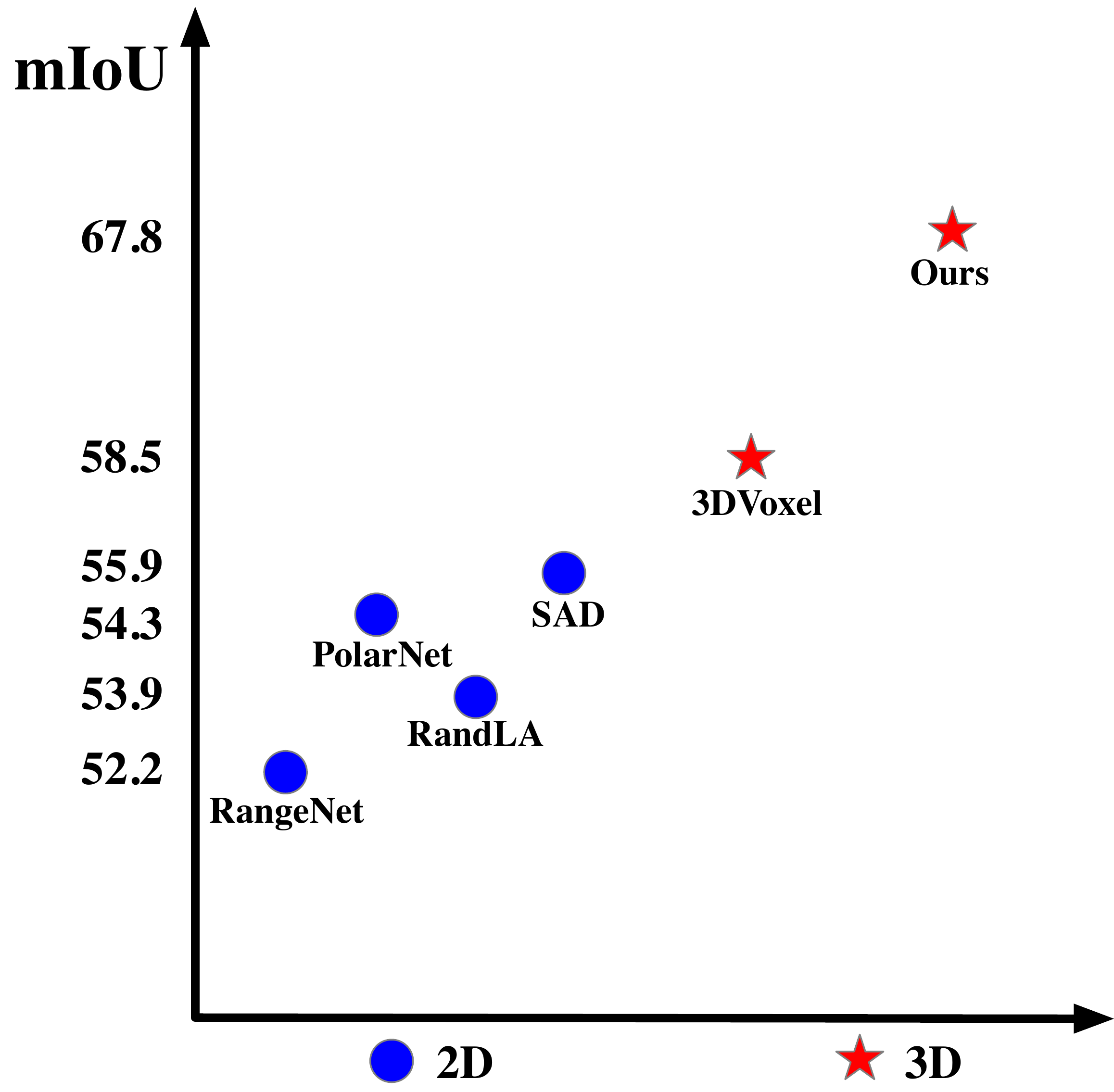}
    \caption{}
    \label{fig:stat1}
    \end{subfigure}
    \vspace{-1ex}
    \caption{(a) Range Image (2D projection) v.s. Cubic Partition v.s. Cylindrical Partition. From top row, it can be found that range image abandons the 3D topology, where 2d convolution processes points in different locations (far away from each other in green circles). From bottom part, cylindrical partition generates the more balanced point distribution than cubic partition (89\% v.s. 61\% cells containing points). (b) Applying the regular 3D voxel partition and 3D convolution directly (\ie, 3DVoxel) gets limited performance gain compared to projection-based (2D) methods~\cite{zhang2020polarnet, hu2020randla, milioto2019rangenet++, xu2020squeezesegv3}, while our method achieves a remarkable performance gain by further tackling the inherent difficulty of outdoor LiDAR point clouds (showing results on SemanticKITTI dataset).
    }
    \label{fig:teaser}
    \vspace{-3ex}
    \end{figure*}
    
    Recently, the advances in deep learning have significantly pushed forward the state of the art in image segmentation. Some existing LiDAR segmentation approaches follow this route to project the 3D point clouds onto a 2D space and process them via 2D convolution networks, including range image based~\cite{milioto2019rangenet++,wu2018squeezeseg} and bird's-eye-view
   image based~\cite{zhang2020polarnet}. However, this group of methods lose and alter the accurate 3D geometric information during the 3D-to-2D projection (as shown in the top row of Fig.~\ref{fig:2d_3d_conv}).
   
   A natural alternative is to utilize the 3D partition and 3D convolution networks to process the point cloud and maintain their 3D geometric relations. However, in our initial attempts, we directly apply the 3D voxelization~\cite{graham20183d, cciccek20163d} and 3D convolution networks to outdoor LiDAR point cloud, only to find very limited performance gain (as shown in Fig.~\ref{fig:stat1}).
   Our investigation into this issue reveals a key difficulty of outdoor LiDAR point cloud, namely sparsity and varying density, which is also the key difference to indoor scenes with dense and uniform-density points. However, previous 3D voxelization methods consider the point cloud as a uniform one and split them via the uniform cube, while neglecting the varying-density property of outdoor point cloud. Consequently, this effect to apply the 3D partition to outdoor point cloud is met with fundamental difficulty.
%   Hence, attention to the sparsity and varying density is crucial to the outdoor point cloud.
   
   Motivated by these findings, we propose a new framework to outdoor LiDAR segmentation that consists of two key components, \ie, 3D cylindrical partition and asymmetrical 3D convolution networks, which maintain the 3D geometric information and handle these issues from partition and networks, respectively. Here, cylindrical partition resorts to the cylinder coordinates to divide the point cloud dynamically according to the distance (Regions that are far away from the origin have much sparse points, thus requiring a larger cell), which produces a more balanced point distribution (as shown in Fig.~\ref{fig:2d_3d_conv}); while asymmetrical 3D convolution networks strengthen the horizontal and vertical kernels to match the point distribution of objects in the driving scene and enhance the robustness to the sparsity. Moreover, voxel based methods might divide the points with different categories into the same cell and cell label encoding would inevitably cause the information loss. To alleviate the interference of lossy label encoding, a point-wise module is introduced to further refine the features obtained from voxel-based network. 
   Overall, the corporation of these components well maintains the geometric relation and tackle the difficulty of outdoor point cloud, thus improving the effectiveness of 3D frameworks.
   
   We evaluate the proposed method on two large-scale outdoor datasets, including SemanticKITTI~\cite{behley2019semantickitti} and nuScenes~\cite{nuscenes}. Our method achieves the \textbf{1st place} in the leaderboard of SemanticKITTI and also outperforms the existing methods on nuScenes with a large margin. We also extend the proposed cylindrical partition and asymmetrical 3D convolution networks to LiDAR panoptic segmentation and LiDAR 3D detection. Experimental results show its strong generalization capability and scalability.
   
   The contributions of this work mainly lie in three aspects:
   (1) We reposition the focus of outdoor LiDAR segmentation from 2D projection to 3D structure, and further investigate the inherent properties (difficulties) of outdoor point cloud. (2) We introduce a new framework to explore the 3D geometric pattern and tackle these difficulties caused by sparsity and varying density, through cylindrical partition and asymmetrical 3D convolution networks. (3) The proposed method achieves the state of art on SemanticKITTI and nuScenes, and also shows strong generalization capability on LiDAR panoptic segmentation and LiDAR 3D detection.

\section{Related Work}
   
   %\noindent In this section, we will introduce prior literature from two aspects.
   
    \begin{figure*}
    \centering
    \includegraphics[width=0.92\linewidth]{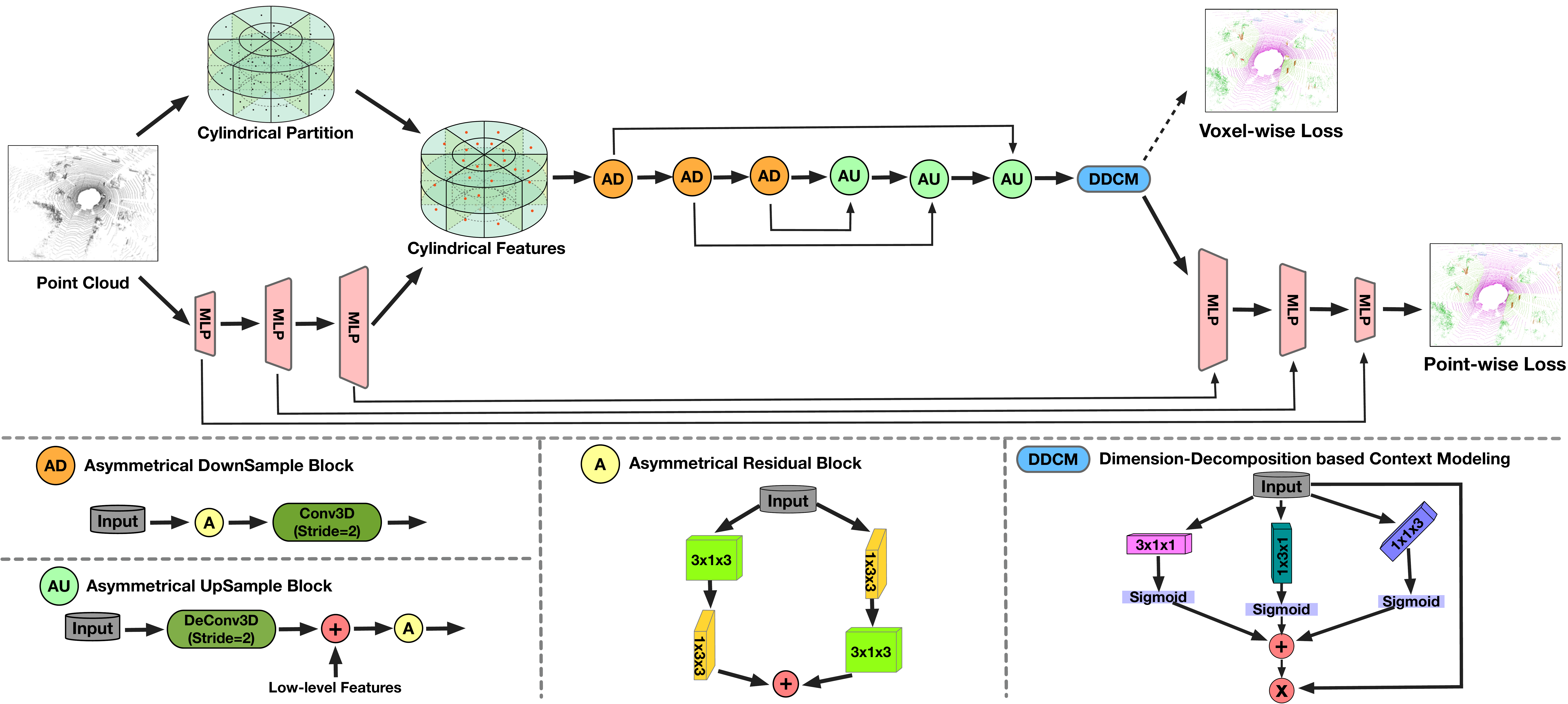}
    \caption{(1): Top half is the overall framework. Here, LiDAR point cloud is fed into MLP to get the point-wise features and then these features are reassigned based on the cylinderical partition. Asymmetrical 3D convolution networks are then used to generate the voxel-wise outputs. Finally, a point-wise module is introduced to refine these outputs. (2): Bottom half elaborates four components, including Asymmetrical Downsample block (AD), Asymmetrical Upsample blcok (AU), Asymmetrical residual block (A) and Dimension-Decomposition based Context Modeling (DDCM).}
    \label{fig:pipeline}
    \end{figure*}
   
   \textbf{Indoor-scene Point Cloud Segmentation.} Indoor-scene point clouds carry out some properties, including generally uniform density, small number of points, and small range of the scene. Mainstream methods~\cite{qi2017pointnet, thomas2019kpconv, wu2019pointconv, wang2019dynamic,velivckovic2017graph, lyu2020learning, engelmann20203d, zhang2020fusion, yan2020pointasnl,wang2019graph,pham2019jsis3d,qi20173d} of indoor point cloud segmentation learn the point features based on the raw point directly, which are often based on the pioneering work, \ie, PointNet, and promote the effectiveness of sampling, grouping and ordering to achieve the better performance. Another group of methods utilize the clustering algorithm~\cite{wang2019dynamic,velivckovic2017graph} to extract the hierarchical point features. However, these methods focusing on indoor point cloud are limited to adapt to the outdoor point cloud under the property of varying density and large range of scenes, and the large number of points also result in the computational difficulties for these methods when deploying from indoor to outdoor. 
%   Hence, most indoor-scene segmentation methods~\cite{qi2017pointnet, thomas2019kpconv, wu2019pointconv, wang2019dynamic,velivckovic2017graph, lyu2020learning, engelmann20203d, zhang2020fusion} often learn the point features from the raw point directly. PointNet~\cite{qi2017pointnet} is the pioneering work to point cloud, which utilized the multi-layer perception and max-pooling to extract features from points. 
%   Moreover, PointNet++~\cite{qi2017pointnet++} further proposed multi-scale sampling to aggregate global and local features. Another group of indoor-scene segmentation~\cite{wang2019dynamic,velivckovic2017graph} utilizes the clustering (including KNN) to extract the point features. 
%   However, these methods are computationally costly and do not take varying sparsity (the property of outdoor-scene LiDAR) into consideration. 

   \textbf{Outdoor-scene Point Cloud Segmentation.}
   Most existing outdoor-scene point cloud segmentation methods~\cite{hu2020randla,cortinhal2020salsanext,milioto2019rangenet++,alonso20203d,zhang12356deep,landrieu2018large} focus on converting the 3D point cloud to 2D grids, to enable the usage of 2D Convolutional Neural Networks. SqueezeSeg~\cite{wu2018squeezeseg}, Darknet~\cite{behley2019semantickitti}, SqueezeSegv2~\cite{wu2019squeezesegv2},
  and RangeNet++~\cite{milioto2019rangenet++} utilize the spherical projection mechanism, which converts the point cloud to 
  a frontal-view image or a range image, and adopt the 2D convolution network on the pseudo image for point cloud segmentation. PolarNet~\cite{zhang2020polarnet} follows the bird's-eye-view projection, which projects point cloud data into bird's-eye-view representation under the polar coordinates. However, these 3D-to-2D projection methods inevitably loss and alter the 3D topology and fails to model the geometric information.
  
  \textbf{3D Voxel Partition.}
  3D voxel partition is another routine of point cloud encoding~\cite{han2020occuseg,tchapmi2017segcloud,graham20183d,cciccek20163d,meng2019vv}. It converts a point cloud into 3D voxels, which mainly retains the 3D geometric information. 
%   3D U-Net~\cite{cciccek20163d} proposes voxel partition and 3D U-Net on biomedical data and shows successful application on difficult microscopic datasets. 
  OccuSeg~\cite{han2020occuseg}, SSCN~\cite{graham20183d} and SEGCloud~\cite{tchapmi2017segcloud} follow this line to utilize the voxel partition and apply regular 3D convolutions for LiDAR segmentation. It is worth noting that while the aforementioned efforts have shown encouraging results, the improvement in the outdoor LiDAR point cloud remains limited. As mentioned above, a common issue is that these methods neglect the inherent properties of outdoor LiDAR point cloud, namely, sparsity and varying density. Compared to these methods, our proposed method resorts to the 3D cylindrical partition and asymmetrical 3D convolution networks to tackle these difficulties.

\textbf{Network Architectures for Segmentation}. Fully Convolutional Network~\cite{long2015fully} is the fundamental work in the deep-learning era. Built upon the FCN, many works aim to improve the performance via exploring the dilated convolution, multi-scale context modeling and attention modeling, including DeepLab\cite{chen2017deeplab, chen2018encoder} and PSP~\cite{zhao2017pyramid}. Recent work utilizes the neural architecture search to find the more effective backbone for the segmentation~\cite{liu2019auto,tang2020searching}. Moreover, U-Net~\cite{ronneberger2015u} proposes a symmetric architecture to incorporate the low-level features. With the great success of U-Net on 2D benchmarks and its good flexibility , many studies for LiDAR segmentation adapt the U-Net to the 3D space~\cite{cciccek20163d}. We also follow this structure to construct our asymmetrical 3D convolution networks.
% However, they often fail to explore the distribution and property of the driving-scene LiDAR point cloud. In this work, two modules, \ie, Asymmetric Residual Block and Dimension-decomposition based Context Modeling, are designed to match the cuboid objects and model the high-rank context information, respectively.

%==============================================================================

\section{Methodology}
\label{sec:methods}

\subsection{Framework Overview}

 %Considering outdoor point clouds covering a large range of urban scenes, our task is to assign the semantic label to each point in the point cloud.
 In the context of semantic segmentation, given a point cloud, our task is to assign the semantic label to each point.
 Based on the comparison between 2D and 3D representation and investigation of the inherent properties of outdoor LiDAR point cloud, we desire to obtain a framework which explores the 3D geometric information and handles the difficulty caused by sparsity and varying-density. To this end, we propose a new outdoor segmentation approach based on the 3D partition and 3D convolution networks. To handle these difficulties of outdoor LiDAR point cloud, namely sparsity and varying density, we first employ the cylindrical partition to generate the more balanced point distribution (more robust to varying density), then apply the asymmetrical 3D convolution networks to power the horizontal and vertical weights, thus well matching the object point distribution in driving scene and enhancing the robustness to the sparsity. 
%  Based on our investigation on the distribution of 2D and 3D point-cloud representations, we discover that the 2D representation obtained from projection would abandon many available 3D structures. To this end, we propose a new outdoor LiDAR segmentation approach based on 3D representation and neural networks. 
%The basic idea is to utilize the geometric information to benefit the Lidar segmentation. 

%As mentioned, these projections compress the 3D distribution and may alter the feature space.
 As shown in the top half of Fig.~\ref{fig:pipeline}, the framework consists of two major components, including cylindrical partition and asymmetrical 3D convolution networks. The LiDAR point cloud is first divided by the cylindrical partition and the features extracted from MLP is then reassigned based on this partition. Asymmetrical 3D convolution networks are then used to generate the voxel-wise outputs. Finally, a point-wise module is introduced to alleviate the interference of lossy cell-label encoding, thus refining the outputs. In the following sections, we will present these components in detail.

\subsection{Cylindrical Partition}

    \begin{figure}[t]
    \centering
    \includegraphics[width=0.82\linewidth]{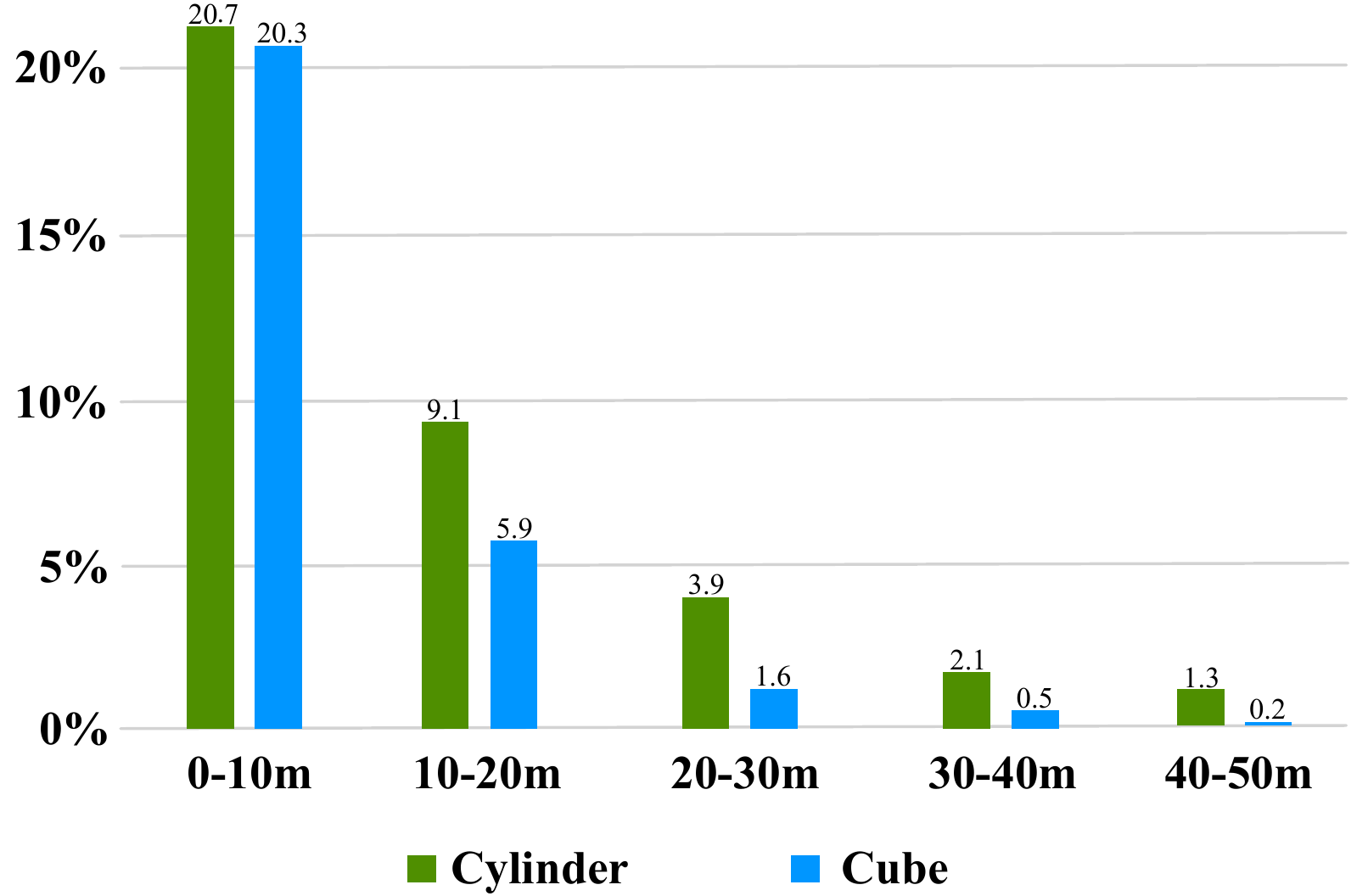}
    \caption{The proportion of non-empty cells at different distances between cylindrical and cubic partition (The results are calculated on the training set of SemanticKITTI). It can be found that cylinder partition makes a higher non-empty proportion and more balanced point distribution, especially for farther-away regions.}
    \label{fig:stat3}
    \end{figure}

As mentioned above, outdoor-scene LiDAR point cloud possesses the property of varying density, where nearby region has much greater density than farther-away region. Therefore, uniform cells splitting the varying-density points would fall into an imbalanced distribution (for example, larger proportion of empty cells). While in the cylinder coordinate system, it utilizes the increasing grid size to cover the farther-away region, and thus more evenly distributes the points across different regions and gives an more balanced representation against the varying density. We perform a statistic to show the proportion of non-empty cells across different distances in Fig.~\ref{fig:stat3}. It can be found that with the distance goes far, cylindrical partition maintains a balanced non-empty proportion due to the increasing grid size while cubic partition suffers the imbalanced distribution, especially in the farther-away regions (about 6 times less than cylindrical partition). Moreover, unlike these projection-based methods project the point to the 2D view, cylindrical partition maintains the 3D grid representation to retain the geometric structure. 

\begin{figure}[t]
    \centering
    \includegraphics[width=1.0\linewidth]{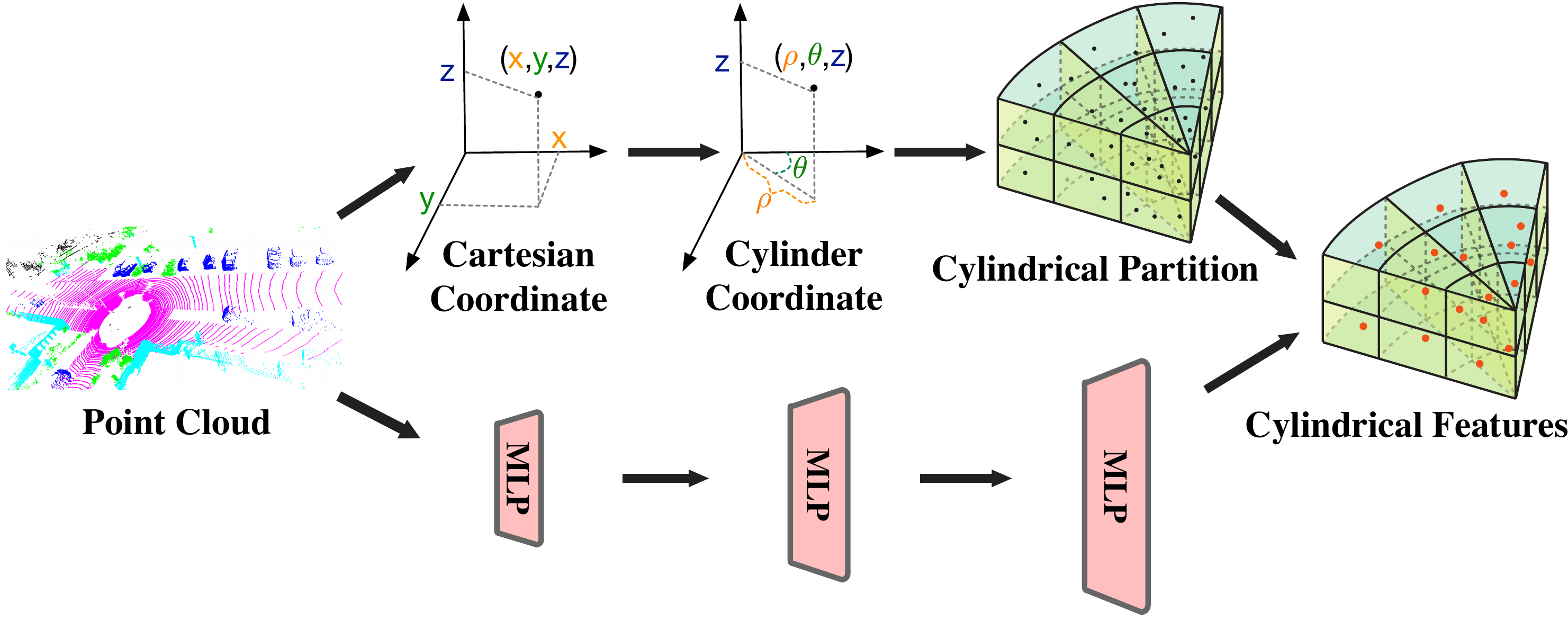}
    \caption{The pipeline of cylindrical partition. We first transform the Cartesian coordinate to Cylinder coordinate and then assign the point-wise features to the structured representation based on the cylindrical partition.}
    \label{fig:cylinder}
    \end{figure}

The workflow is illustrated in Fig.~\ref{fig:cylinder}. We first transform the points on Cartesian coordinate system to the Cylinder coordinate system. This step transforms the points ($x, y, z$) to points ($\rho, \theta, z$), where radius $\rho$ (distance to origin in x-y axis) and azimuth $\theta$ (angle from x-axis to y-axis) are calculated. Then cylindrical partition performs the split on these three dimensions, note that in the cylinder coordinate, more farther-away region, larger cell. 
Point-wise features obtained from the MLP are reassigned based on the result of this partition to get the cylindrical features.
After these steps, we unroll the cylinder from 0-degree and get the 3D cylindrical representation $\mathbb{R}\in C \times H\times W \times L$, where $C$ denotes the feature dimension and $H, W, L$ mean the radius, azimuth and height.
Subsequent asymmetrical 3D convolution networks will be performing on this representation.

\subsection{Asymmetrical 3D Convolution Network}

Since the driving-scene point cloud carries out the specific object shape distribution, including car, truck, bus, motorcycle and other cubic objects, we aim to follow this observation to enhance the representational power of a standard 3D convolution. Moreover, recent literature~\cite{wang2019shape,ding2019acnet} also shows that the central crisscross weights count more in the square convolution kernel. In this way, we devise the asymmetrical residual block to strengthen the horizontal and vertical responses and match the object point distribution. 
Based on the proposed asymmetrical residual block, we further build the asymmetrical downsample block and asymmetrical upsample block to perform the downsample and upsample operation. Moreover, a dimension-decomposition based context modeling (termed as DDCM) is introduced to explore the high-rank global context in decomposite-aggregate strategy. We detail these components in the bottom of Fig.~\ref{fig:pipeline}

\paragraph{Asymmetrical Residual Block}

\begin{figure}[t]
    \centering
    \includegraphics[width=0.88\linewidth]{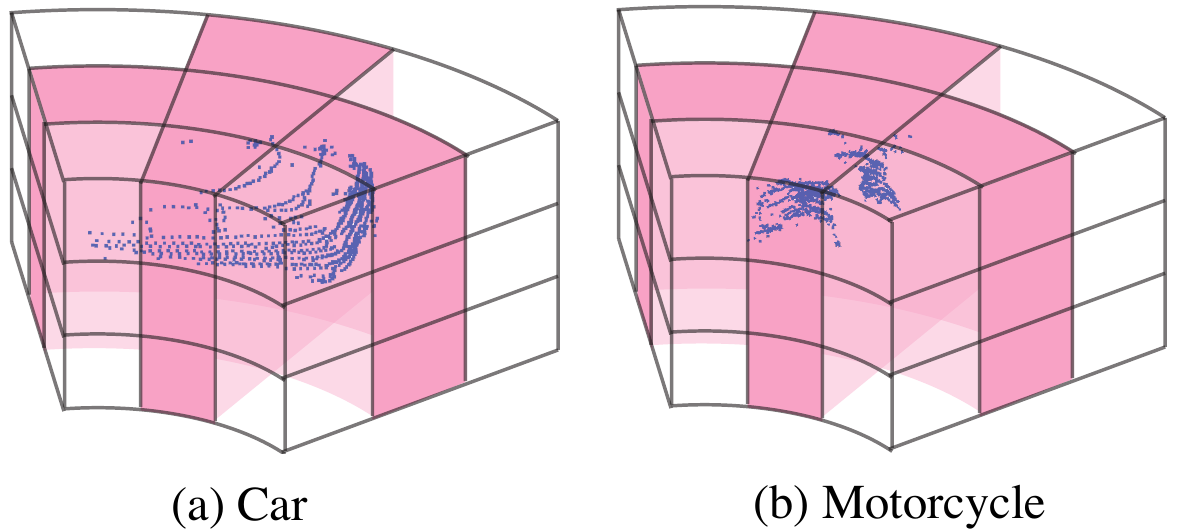}
    \caption{An illustration of asymmetrical residual block, where two asymmetrical kernels are stacked to power the skeleton. It can be observed that asymmetrical residual block focuses on the horizontal and vertical kernels.}
    \label{fig:asym}
\end{figure}

Motivated by the observation and conclusion in~\cite{wang2019shape,ding2019acnet}, the asymmetrical residual block strengthens the horizontal and vertical kernels, which matches the point distribution of object in the driving scene and explicitly makes the skeleton of the kernel powerful, thus enhancing the robustness to the sparsity of outdoor LiDAR point cloud. We use the Car and Motorcycle as the example to show the asymmetrical residual block in Fig.~\ref{fig:asym}, where 3D convolutions are performing on the cylindrical grids. Moreover, the proposed asymmetrical residual block also saves the computation and memory cost compared to the regular square-kernel 3D convolution block. By incorporating the asymmetrical residual block, the asymmetrical downsample block and upsample block are designed and our asymmetrical 3D convolution networks are built via stacking these downsample and upsample blocks.

\paragraph{Dimension-Decomposition based Context Modeling}
Since the global context features should be high-rank to have enough capacity to capture the large context varieties~\cite{zhang2019co}, it is hard to construct these features directly. We follow the tensor decomposition theory~\cite{chen2020tensor} to build the high-rank context as a combination of low-rank tensors, where we use three rank-1 kernels to obtain the low-rank features and then aggregate them together to get the final global context.

\subsection{Point-wise Refinement Module}

\begin{figure}[t]
    \centering
    \includegraphics[width=0.76\linewidth]{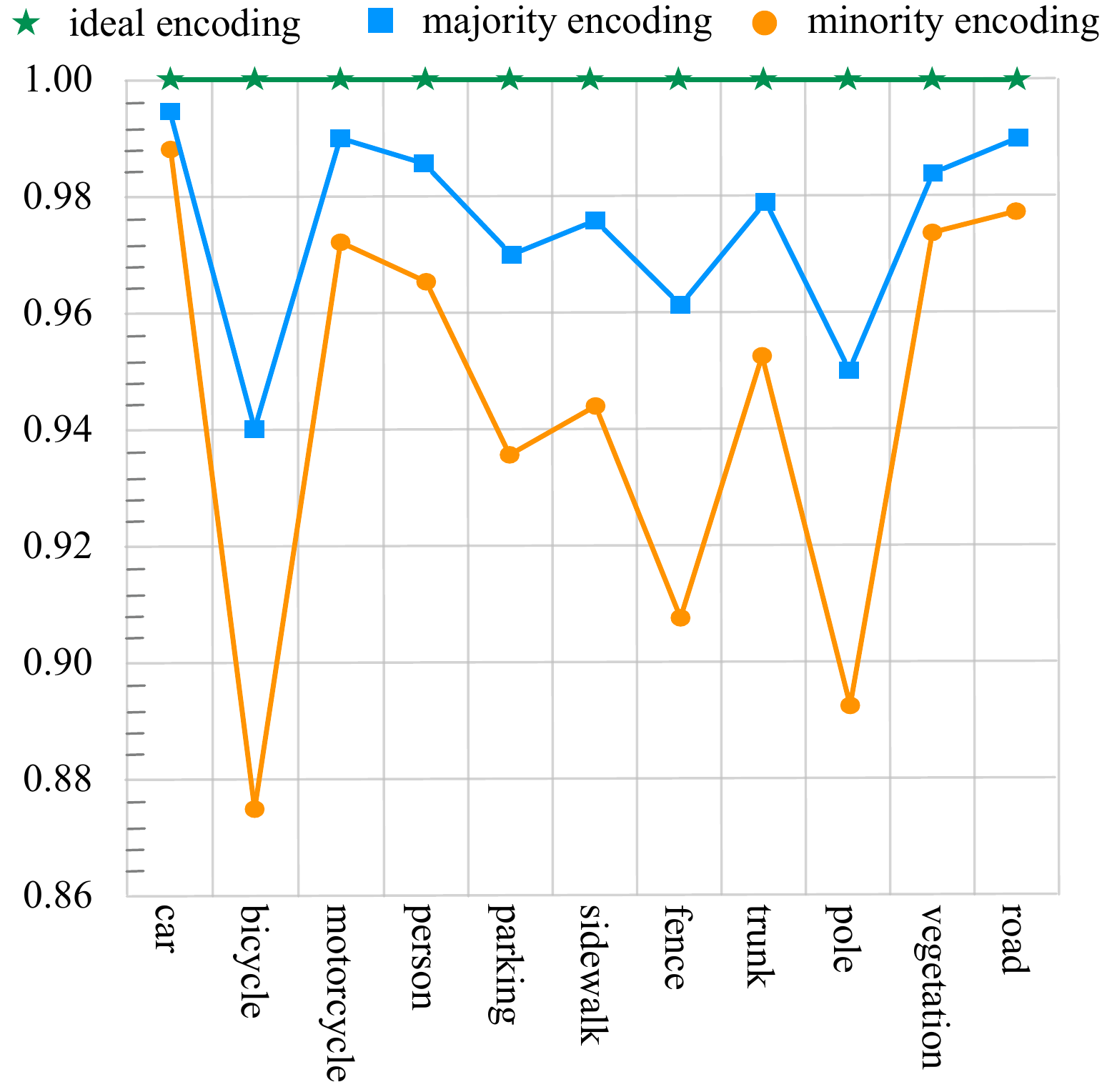}
    \caption{Upper bound of mIoU with different label encoding methods (\ie, majority and minority encoding). It can be found that no matter what encoding methods are, the information loss always occurs, which is also the reason for point-wise refinement. }
    \label{fig:stat2}
\end{figure}

Partition-based methods predict one label for each cell. Although partition-based methods effectively explore the large-range point cloud, however, this group of method, including cube-based and cylinder-based, inevitably suffers from the lossy cell-label encoding, \eg, points with different categories are divided into same cell, and this case would cause the information loss. We make a statistic to show the effect of different label encoding methods in Fig.~\ref{fig:stat2}, where majority encoding means using the major category of points inside a cell as the cell label and minority encoding indicates using the minor category as the cell label. It can be observed that both of them cannot reach the 100 percent mIoU (ideal encoding) and inevitably have the information loss. Here, the point-wise refinement module is introduced to alleviate the interference of lossy cell-label encoding. We first project the voxel-wise features to the point-wise based on the point-voxel mapping table. Then the point-wise module takes both point features before and after 3D convolution networks as the input, and fuses them together to refine the output.

\subsection{Objective Function}

The total objective of our method consists of two components, including voxel-wise loss and point-wise loss. It can be formulated as $\mathbb{L} = \mathbb{L}_{voxel} + \mathbb{L}_{point}$.
For the voxel-wise loss ($\mathbb{L}_{voxel}$), we follow the existing methods~\cite{cortinhal2020salsanext,hu2020randla} and use the weighted cross-entropy loss and lovasz-softmax~\cite{berman2018lovasz} loss to maximize the point accuracy and the intersection-over-union score, respectively. For point-wise loss ($\mathbb{L}_{point}$), we only use the weighted cross-entropy loss to supervise the training. During inference, the output from point-wise refinement module is used as the final output. For the optimizer, Adam with an initial learning rate of $1e^{-3}$ is employed. 
% Total optimization has 40 epochs with the batch size = 2.

\section{Experiments}

In this section, we first provide the detailed experimental setup, then evaluate the proposed method on two large-scale datasets, \ie, SemanticKITTI and nuScenes. Furthermore, extensive ablation studies are conducted to validate each component. In the end, we extend our method to LiDAR panoptic segmentation and 3D detection to verify its scalability and generalization ability. 
% More experimental results (including the results on A2D2 dataset~\cite{geyer2019a2d2}) and visualization are shown in the supplementary material. 

\begin{table*}[t]
\caption{Results of our proposed method and state-of-the-art LiDAR Segmentation methods on SemanticKITTI test set. Note that all results are obtained from the literature, where post-processing, flip \& rotation test ensemble, \etc are also applied.}
\label{semantickitti}
\centering
\begin{adjustbox}{width=\textwidth}
\begin{tabular}{c|c|c|c|c|c|c|c|c|c|c|c|c|c|c|c|c|c|c|c|c}
\hline
\textbf{Methods} & \textbf{mIoU} & \rotatebox{90}{car} &  \rotatebox{90}{bicycle} & \rotatebox{90}{motorcycle} & \rotatebox{90}{truck} & \rotatebox{90}{other-vehicle} & \rotatebox{90}{person} & \rotatebox{90}{bicyclist} & \rotatebox{90}{motorcyclist} & \rotatebox{90}{road} & \rotatebox{90}{parking} & \rotatebox{90}{sidewalk} & \rotatebox{90}{other-ground} &
\rotatebox{90}{building} & \rotatebox{90}{fence} & \rotatebox{90}{vegetation} & \rotatebox{90}{trunk} & \rotatebox{90}{terrain} & \rotatebox{90}{pole} & \rotatebox{90}{traffic} \\
\hline
\hline
% PointNet~\cite{qi2017pointnet} & 146 & 463 & 013 & 003 &  001 &  008 &  002 &  002 & 0 &  616 &  158 &  357 &  014 & 414 & 12.9 & 310 & 046 & 176 & 024 & 037  \\
% \hline
% Splatnet~\cite{su2018splatnet} & 228 & 666 & 0 & 0 & 0 & 0 & 0 & 0 & 0 & 704 & 008 & 415 & 0 & 687 & 278 & 723 & 359 & 358 & 138 & 0\\
% \hline
TangentConv~\cite{tatarchenko2018tangent} & 35.9 & 86.8 & 1.3 & 12.7 & 11.6 & 10.2 & 17.1 & 20.2 & 0.5 & 82.9 & 15.2 & 61.7 & 9.0 & 82.8 & 44.2 & 75.5 & 42.5 & 55.5 & 30.2 & 22.2\\
\hline
Darknet53~\cite{behley2019semantickitti} & 49.9 & 86.4 & 24.5 & 32.7 & 25.5 & 22.6 & 36.2 & 33.6 & 4.7 & 91.8 & 64.8 & 74.6 & {27.9} & 84.1 & 55.0 & 78.3 & 50.1 & 64.0 & 38.9 & 52.2 \\
\hline
RandLA-Net~\cite{hu2020randla} & 50.3 & 94.0 & 19.8 & 21.4 & {42.7} & 38.7 & 47.5 & 48.8 & 4.6  & 90.4 & 56.9 & 67.9 & 15.5 & 81.1 & 49.7 & 78.3 & 60.3 & 59.0 & 44.2 & 38.1 \\
\hline
RangeNet++~\cite{milioto2019rangenet++} & 52.2 & 91.4 & 25.7 & 34.4 & 25.7 & 23.0 & 38.3 &  38.8 & 4.8 & {91.8} & {65.0} & 75.2 & 27.8 & 87.4 & 58.6 & 80.5 & 55.1 & 64.6 & 47.9 & 55.9 \\
\hline
PolarNet~\cite{zhang2020polarnet} & 54.3 & 93.8 & 40.3 & 30.1 & 22.9 & 28.5 & 43.2 & 40.2 & 5.6 & 90.8 & 61.7 & 74.4 & 21.7 & {90.0} & 61.3 & 84.0 & 65.5 & 67.8 & 51.8 & 57.5  \\
\hline
SqueezeSegv3~\cite{xu2020squeezesegv3} & 55.9 & 92.5 & 38.7 & 36.5 & 29.6 & 33.0 & 45.6 & 46.2 & {20.1} & 91.7 & 63.4 & 74.8 & 26.4 & 89.0 & 59.4 & 82.0 & 58.7 & 65.4 & 49.6 & 58.9  \\
\hline
Salsanext~\cite{cortinhal2020salsanext} & 59.5 & 91.9 & 48.3 & 38.6 & 38.9 & 31.9 & 60.2 & 59.0 & 19.4 & 91.7 & 63.7 & 75.8 & 29.1 & 90.2 & 64.2 & 81.8 & 63.6 & 66.5 & 54.3 & 62.1 \\
\hline
KPConv~\cite{thomas2019kpconv} &58.8& 96.0&32.0 & 42.5 & 33.4&44.3&61.5 & 61.6 & 11.8 & 88.8 & 61.3&  72.7&31.6& \bf{95.0} & 64.2 & 84.8 & 69.2 & 69.1 & 56.4 & 47.4 \\
\hline
FusionNet~\cite{zhang12356deep} & 61.3 & 95.3 & 47.5 & 37.7 & 41.8 & 34.5 & 59.5 & 56.8 & 11.9 & 91.8 & 68.8 & 77.1 & 30.8 & 92.5 & \bf{69.4} & 84.5 & 69.8 & 68.5&60.4 & \bf{66.5} \\ 
\hline
KPRNet~\cite{kochanov2020kprnet} & 63.1 & 95.5&54.1& 47.9&23.6 & 42.6&65.9 & 65.0 & 16.5 & \bf{93.2} & \bf{73.9} & \bf{80.6} & 30.2 & 91.7 & {68.4} & \bf{85.7} & 69.8 & 71.2 & 58.7 & 64.1 \\
\hline
TORANDONet~\cite{gerdzhev2020tornado} & 63.1 &94.2& 55.7& 48.1& 40.0& 38.2& 63.6& 60.1& 34.9& 89.7& 66.3& 74.5& 28.7& 91.3& 65.6& 85.6& 67.0& \bf{71.5} & 58.0 & {65.9} \\
\hline
SPVNAS~\cite{tang2020searching} & 66.4 & - & - & - & - & - & - & - & - & - & - & - & - & - & - & - & - & - & - & - \\
\hline
\hline
Ours & \bf{67.8} & \bf{97.1} & \bf{67.6} & \bf{64.0} & \textbf{59.0} & \bf{58.6} & \bf{73.9} & \bf{67.9} & \bf{36.0} & {91.4} & {65.1} & {75.5} & \bf{32.3} & {91.0} & {66.5} & {85.4} & \bf{71.8} & {68.5} & \bf{62.6} & {65.6}  \\
 \hline
\end{tabular}
\end{adjustbox}
\end{table*}

\begin{table*}[t]
\caption{Results of our proposed method and other LiDAR Segmentation methods on nuScenes validation set.}
\label{nuscenes}
\centering
\begin{adjustbox}{width=\textwidth}
\begin{tabular}{c|c|c|c|c|c|c|c|c|c|c|c|c|c|c|c|c|c}
\hline
\textbf{Methods} & \textbf{mIoU} & \rotatebox{90}{barrier} &  \rotatebox{90}{bicycle} & \rotatebox{90}{bus} & \rotatebox{90}{car} & \rotatebox{90}{construction} & \rotatebox{90}{motorcycle} & \rotatebox{90}{pedestrian} & \rotatebox{90}{traffic-cone} & \rotatebox{90}{trailer} & \rotatebox{90}{truck} & \rotatebox{90}{driveable} & \rotatebox{90}{other} &
\rotatebox{90}{sidewalk} & \rotatebox{90}{terrain} & \rotatebox{90}{manmade} & \rotatebox{90}{vegetation} \\
\hline
\hline
RangeNet++~\cite{milioto2019rangenet++} & 65.5 & 66.0 & 21.3 & 77.2 & 80.9 & 30.2 & 66.8 & 69.6 &  52.1 & 54.2 & {72.3} & {94.1} & 66.6 & 63.5 & 70.1 & 83.1 & 79.8 \\
\hline
PolarNet~\cite{zhang2020polarnet} & 71.0 & 74.7 & 28.2 & 85.3 & 90.9 & 35.1 & 77.5 & 71.3 & 58.8 & 57.4 & 76.1 & 96.5 & 71.1 & 74.7 & {74.0} & 87.3 & 85.7  \\
\hline
Salsanext~\cite{cortinhal2020salsanext} & 72.2 & 74.8 & 34.1 & 85.9 & 88.4 & 42.2 & 72.4 & 72.2 & 63.1 & 61.3 & 76.5 & 96.0 & 70.8 & 71.2 & 71.5 & 86.7 & 84.4 \\
\hline
\hline
Ours & \bf{76.1} & \bf{76.4} & \bf{40.3} & \bf{91.2} & \bf{93.8} & \textbf{51.3} & \bf{78.0} & \bf{78.9} & \bf{64.9} & \bf{62.1} & \bf{84.4} & \bf{96.8} & \bf{71.6} & \bf{76.4} & \bf{75.4} & \bf{90.5} & \bf{87.4}  \\
 \hline
\end{tabular}
\end{adjustbox}
\end{table*}

\subsection{Dataset and Metric}

\vspace{0.5ex}
\noindent\textbf{SemanticKITTI~\cite{behley2019semantickitti}}~~~ is a large-scale driving-scene dataset for point cloud segmentation, including semantic segmentation and panoptic segmentation. It is derived from the KITTI Vision Odometry Benchmark and collected in Germany with the Velodyne-HDLE64 LiDAR. The dataset consists of 22 sequences, splitting sequences 00 to 10 as training set (where sequence 08 is used as the validation set), and sequences 11 to 21 as test set. 19 classes are remained for training and evaluation after merging classes with different moving status and ignore classes with very few points. 

\vspace{0.5ex}
\noindent\textbf{nuScenes~\cite{nuscenes}}~~~ It collects 1000 scenes of 20s duration with 32 beams LiDAR sensor. The number of total frames is 40,000, which is sampled at 20Hz. They also officially split the data into training and validation set. After merging similar classes and removing rare classes, total 16 classes for the LiDAR semantic segmentation are remained. 

For both two datasets, cylindrical partition splits these point clouds into 3D representation with the size = $480\times360\times32$, where three dimensions indicate the radius, angle and height, respectively.

\vspace{0.5ex}
\noindent\textbf{Evaluation Metric}~~~
To evaluate the proposed method, we follow the official guidance to leverage mean intersection-over-union (mIoU) as the evaluation metric defined in~\cite{behley2019semantickitti,nuscenes}, which can be formulated as:
$
IoU_{i} = \frac{TP_{i}}{TP_{i} + FP_{i} + FN_{i}}
$
where $TP_{i}, FP_{i}, FN_{i}$ represent true positive, false positive, and false negative predictions for
class $i$ and the mIoU is the mean value of $IoU_{i}$ over all classes.

\subsection{Results on SemanticKITTI}

In this experiment, we compare the results of our proposed method with existing state-of-the-art LiDAR segmentation methods on SemanticKITTI test set. As shown in Table~\ref{semantickitti}, our method outperforms all existing methods in term of mIoU. Compared to the projection-based methods on 2D space, including Darknet53~\cite{behley2019semantickitti}, SqueezeSegv3~\cite{xu2020squeezesegv3}, RangeNet++~\cite{milioto2019rangenet++} and PolarNet~\cite{zhang2020polarnet}, our method achieves 8\% $\thicksim$ 17\% performance gain in term of mIoU due to the modeling of 3D geometric information. Compared to some voxel partition and 3D convolution based methods, including FusionNet~\cite{zhang12356deep}, TORANDONet~\cite{gerdzhev2020tornado} (multi-view fusion based method) and SPVNAS~\cite{tang2020searching} (utilizing the neural architecture search for LiDAR segmentation), the proposed method also performs better than these 3D convolution based methods, where the cylindrical partition and asymmetrical 3D convolution networks well handle the difficulty of driving-scene LiDAR point cloud that is neglected by these methods.

\subsection{Results on nuScenes}

Since nuScenes dataset does not release the evaluation server for LiDAR segmentation, we report the results on the validation set, where RangeNet++ and Salsanext perform the post-processing. As shown in Table~\ref{nuscenes}, our method achieves better performance than existing methods in all categories. Specifically, the proposed method obtains about 4\% $\thicksim$ 7\% performance gain than projection-based methods. Moreover, for these categories with sparse points, such as bicycle and pedestrian, our method significantly outperforms existing approaches, which also demonstrates the effectiveness of the proposed method to tackle the sparsity and varying density.

\subsection{Ablation Studies}

In this section, we perform the thorough ablation experiments to investigate the effect of different components in our method. We also design several variants of asymmetrical residual block to verify our claim that strengthening the horizontal and vertical kernels power the representation ability for driving-scene point cloud.

\begin{table}[t]
\caption{Ablation studies for network components on SemanticKITTI validation set. PR denotes the point-wise refinement module.}
\label{table_net_components}
\centering
% \small
\setlength{\tabcolsep}{4.2pt}
%\begin{adjustbox}{width=\textwidth}
\begin{tabular*}{1.0\linewidth}{c c c c c | c}
\hline
{Baseline} & Cylinder & {Asym-CNN} & {DDCM} & {PR} & {mIoU} \\
% &  & \textbf{based Context Modeling} &  &  \\
\hline
\hline
{\cmark} & & & & & 58.1 \\
%\hline
\cmark & \cmark & & & & 61.0 \\
%\hline
\cmark & \cmark & \cmark & & & 63.8\\
%\hline
\cmark & \cmark & \cmark & \cmark & & 65.2 \\
\cmark & \cmark & \cmark & \cmark & \cmark & 65.9 \\
\hline
\end{tabular*}
%\end{adjustbox}
\end{table}

\vspace{1ex}
\noindent\textbf{Effects of Network Components}~~~
In this part, we make several variants of our model to validate the contributions of different components. The results on SemanticKITTI validation set are reported in Table~\ref{table_net_components}. Baseline method denotes the framework using 3D voxel partition and 3D convolution networks. It can be observed that cylindrical partition performs much better than cubic-based partition with about 3\% mIoU gain and asymmetrical 3D convolution networks also significantly boost the performance about 3\% improvement, which demonstrates that both cylindrical partition and asymmetrical 3D convolution networks are crucial in the proposed method. Furthermore, dimension-decomposition based context modeling delivers the effective global context features, which yields an improvement of 1.4\%. Point-wise refinement module further pushes forward the performance based on the strong model, about 0.7\%.

\begin{figure}[t]
    \centering
    \includegraphics[width=1.0\linewidth]{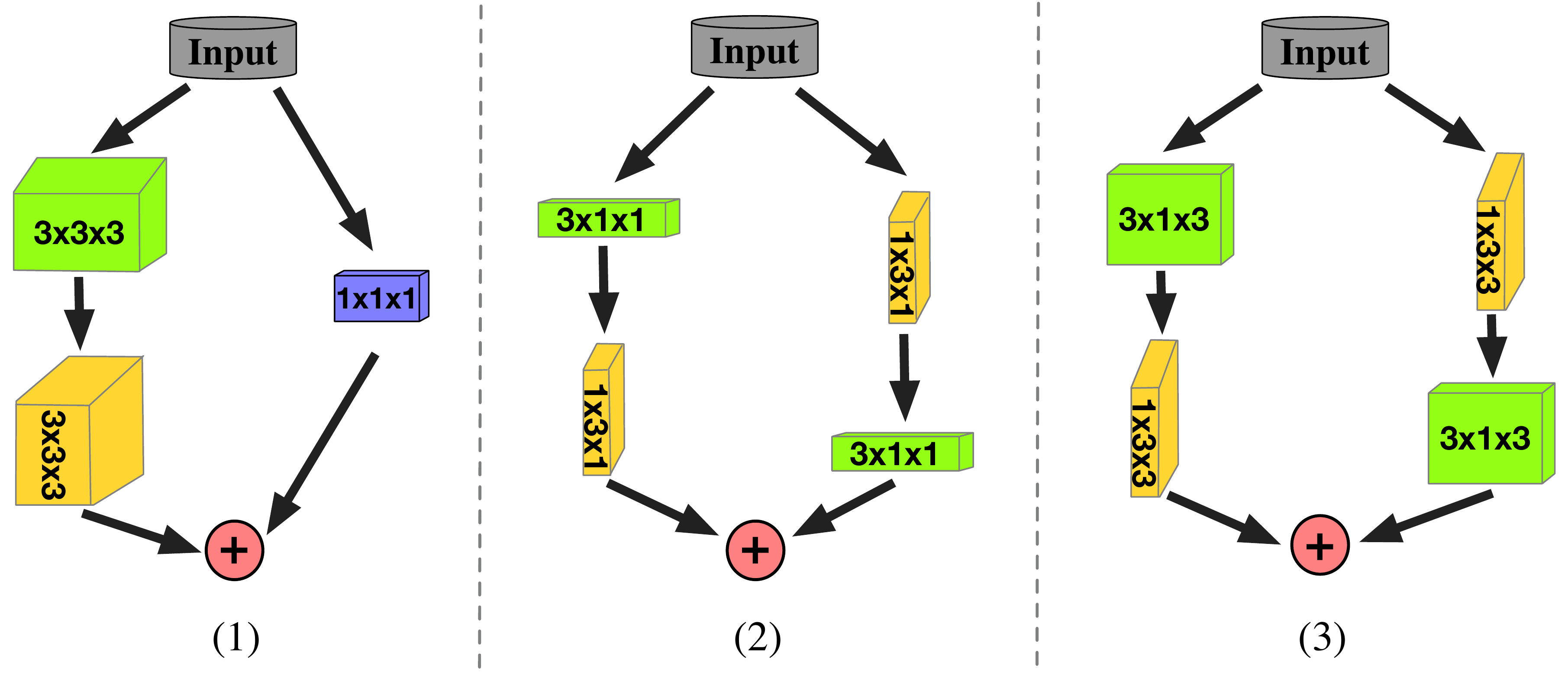}
    \caption{We design three blocks for ablation studies of asymmetrical residual block, including (1) regular residual block , (2) 1D-asymmetrical residual block without height and (3) the proposed asymmetrical residual block.}
    \label{fig:asym_vari}
\end{figure}

\vspace{1ex}
\noindent\textbf{Variants of Asymmetrical Residual Block}~~~
To verify the effectiveness of the proposed asymmetrical residual block, we design several variants of asymmetrical residual block to investigate the effect of horizontal and vertical kernel enhancement (as shown in Fig.~\ref{fig:asym_vari}). The first variant is the regular residual block without any asymmetrical structure. The second one is the 1D-asymmetrical residual block, which utilizes the 1D asymmetrical kernels without height and also strengthens the horizontal and vertical kernels in one-dimension.

We conduct the ablation studies on SemanticKITTI validation set. Note that we use the cylindrical partition as the partition method and stack these proposed variants to build the 3D convolution networks. We report the results in Table~\ref{tab:asym_vari}. It can be found that although the 1D-Asymmetrical residual block only powers the horizontal and vertical kernels in one-dimension without considering the height, it still achieves 1.3\% gain in term of mIoU, which demonstrates the effectiveness of horizontal and vertical kernel strengthening. After taking the height into the consideration, the proposed asymmetrical residual block further matches the object distribution in driving scene and powers the skeleton of kernels to enhance the robustness to the sparsity. From Table~\ref{tab:asym_vari}, the proposed asymmetrical residual block significantly boosts the performance with about 3\% improvements, where large improvement can be observed on some instance categories (about 10\% gain), including truck, person and motorcycle, because it matches the point distribution of object and enhances the representational power.

\begin{table}
%\vspace{-9ex}
%\small
\caption{Ablation studies for asymmetrical residual block.}
\centering
%\vspace{2ex}
\begin{tabular*}{0.75\linewidth}{l|c}
\hline
Methods  & mIoU \\
\hline
\hline
Regular residual block & 61.0\\
\hline
1D-Asymmetrical residual block & 62.0 \\
\hline
Asymmetrical residual block & 63.8 \\
\hline
\end{tabular*}
% \vspace{1ex}
\label{tab:asym_vari}
\end{table}

\subsection{Generalization Analyses}

\begin{table*}[ht]
\caption{LiDAR panoptic segmentation results on the validation set of SemanticKITTI.}
\vspace{-3ex}
    \begin{center}
    \small{
        \begin{tabular}{l|c|ccc|ccc|ccc|c}
            \hline
            Method & \textbf{\PQ} & \PQda & \RQ & \SQ & \PQth & \RQth & \SQth & \PQst & \RQst & \SQst & \miou \\
            \hline\hline
            KPConv \cite{thomas2019kpconv} +
            PV-RCNN \cite{shi2020pv}              & 51.7         & 57.4          & 63.1          & \textbf{78.9} & 46.8          & 56.8          & \textbf{81.5} & \textbf{55.2} & \textbf{67.8} & \textbf{77.1} & 63.1          \\
            PointGroup \cite{jiang2020pointgroup} & 46.1         & 54.0          & 56.6          & 74.6          & 47.7          & 55.9          & 73.8          & 45.0          & 57.1          & 75.1                      & 55.7          \\
            LPASD \cite{milioto2020iros}             & 36.5 & 46.1 & -    & -    & -    & 28.2 & -    & -    & -    & -    & 50.7 \\
            % PanosterK\cite{2020arXiv201015157G}  & 55.6 & -    & 66.8 & 79.9 & 56.6 & 65.8 & -    & -    & -    & -    & 61.1 \\
            \hline
            % Our Baseline                         & 56.4          & 62.0          & 67.1          & 76.5          & 58.6          & 66.7          & 75.8          & 54.8          & 67.4          & 77.0                      & \textbf{63.5} \\
            \hline
            Ours                          & \textbf{56.4} & \textbf{62} & \textbf{67.1} & 76.5          & \textbf{58.8} & \textbf{66.8} & 75.8          & 54.8          & 67.4          & \textbf{77.1} & \textbf{63.5} \\
            \hline
        \end{tabular}
    }
    \end{center}
    % \vspace{-0.6cm}
    \label{tab:semkitti_val}
\end{table*}

\begin{table}[t]
    \caption{LiDAR 3D detection results on nuScenes dataset. CyAs denotes the {\bf{Cy}}lindrical partition and {\bf{As}}ymmetrical 3D convolution networks.}
            % \small
            %\setlength{\tabcolsep}{7.0pt}
            % \vspace{-2ex}
            \centering
            %%\vspace{1ex}
            \begin{tabular*}{0.68\linewidth}{l|c|c}
            \hline
            Methods  & mAP & NDS \\
            \hline
            \hline
            PointPillar~\cite{lang2019pointpillars} & 30.5 & 45.3 \\
            \hline
            PP + Reconfig~\cite{wang2020reconfigurable} & 32.5 & 50.6 \\
            \hline
            SECOND~\cite{yan2018second} & 31.6 & 46.8\\
            \hline
            SECOND + CyAs & 36.4 & 51.7\\
            \hline
            SSN~\cite{zhu2020ssn}&  46.3 & 56.9 \\
            \hline
            SSN + CyAs & 47.7 & 58.2 \\
            \hline
            \end{tabular*}
    \label{tab:gene_nusc}
\end{table}

In this section, we extend the proposed cylindrical partition and asymmetrical 3D convolution networks to other LiDAR-based tasks, including LiDAR panoptic segmentation and LiDAR 3D detection. These experimental results demonstrate the generalization ability of our proposed method and its effectiveness on LiDAR point cloud processing.
In what follows, we will describe these two tasks and present how we adapt the proposed model in details.

\vspace{1ex}
\noindent\textbf{Generalize to LiDAR Panoptic Segmentation}~~~
Panoptic segmentation is first proposed in \cite{kirillov2019panoptic} as a new task, in which semantic segmentation is performed for background classes and instance segmentation for foreground classes and these two groups of category are also termed as {\bf{stuff}} and {\bf{things}} classes, respectively.
Behley \etal \cite{behley2020benchmark} extend the task to LiDAR point clouds and propose the LiDAR panoptic segmentation. In this experiment, we conduct the panoptic segmentation on SemanticKITTI dataset and report results on the validation set. For the evaluation metrics, we follow the metrics defined in \cite{behley2020benchmark}, where they are the same as that of image panoptic segmentation defined in \cite{kirillov2019panoptic} including Panoptic Quality (PQ), Segmentation Quality (SQ) and Recognition Quality (RQ) which are calculated across all classes. \PQda{} is defined by swapping \PQ{} of each {\bf{stuff}} class to its IoU and averaging over all classes like \PQ{} does. Since the categories in panoptic segmentation contain two groups, \ie, {\bf{stuff}} and {\bf{things}}, these metrics are also performed separately on these two groups, including PQ\textsuperscript{Th}, PQ\textsuperscript{St}, RQ\textsuperscript{Th}, RQ\textsuperscript{St}, SQ\textsuperscript{Th} and SQ\textsuperscript{St}, where Panoptic Quality (PQ) is usually used as the first criteria.

In this experiment, we use the proposed cylindrical partition as the partition method and asymmetrical 3D convolution networks as the backbone. Moreover, a semantic branch is used to output the semantic labels for stuff categories, and an instance branch is introduced to generate the instance-level features and further extract their instance IDs for things categories through heuristic clustering algorithms (we use mean-shift in the implementation).

We report the results in Table~\ref{tab:semkitti_val}. It can be found that our method achieves much better performance than existing methods~\cite{milioto2020iros,jiang2020pointgroup}. In terms of PQ, we have about 4.7\% point improvement, and particularly for the thing categories, our method significantly outperforms state-of-the-art in terms of PQ\textsuperscript{Th} and PQ\textsuperscript{St} with a large margin of 10\% points. Experimental results demonstrate the effectiveness of the proposed method and its good generalization ability.

\vspace{1ex}
\noindent\textbf{Generalize to LiDAR 3D Detection}~~~ LiDAR 3D detection aims to localize and classify the multi-class objects in the point cloud. SECOND~\cite{yan2018second} first utilizes the 3D voxelization and 3D convolution networks to perform the single-stage 3D detection. In this experiment, we follow SECOND method and replace the regular voxelization and 3D convolution with the proposed cylindrical partition and asymmetrical 3D convolution networks, respectively. Similarly, to verify its scalability, we also extend the proposed modules to SSN~\cite{zhu2020ssn}. 
The experiments are conducted on nuScenes dataset and the cylindrical partition also generates the $480\times360\times32$ representation. For the evaluation metrics, we follow the official metrics defined in nuScenes, \ie, mean average precision (mAP) and nuScenes detection score (NDS). 

The results are shown in Table~\ref{tab:gene_nusc}. PP + Reconfig~\cite{wang2020reconfigurable} is a partition enhancement approach based on PointPillar~\cite{lang2019pointpillars}, while our SECOND + CyAs performs better with similar backbone, which indicates the superiority of the cylindrical partition.
 We then extend the proposed method (\ie, CyAs) to two baseline methods, termed as SECOND + CyAs and SSN + CyAs, respectively.
 By comparing these two models with their extensions, it can be observed that the proposed Cylindrical partition and Asymmetrical 3D convolution networks boost the performance consistently, even for the strong baseline \ie,~SSN, which demonstrates the effectiveness and scalability of our model.

\section{Conclusion}

In this paper, we have proposed a cylindrical and asymmetrical 3D convolution networks for LiDAR segmentation, where it maintains the 3D geometric relation. Specifically, two key components, the cylinder partition and asymmetrical 3D convolution networks, are designed to handle the inherent difficulties in outdoor LiDAR point cloud, namely sparsity and varying density, effectively and robustly.
We conduct the extensive experiments and ablation studies, where the model achieves the {{1st place}} in SemanticKITTI and state-of-the-art in nuScenes, and keeps good generalization ability to other LiDAR based tasks, including LiDAR panoptic segmentation and LiDAR 3D detection.

%===============================================================================

% \input{doc/method}

% \input{doc/exper}

% \input{doc/conclusion}

{\small
\bibliographystyle{ieee_fullname}
\bibliography{egbib}
}

\end{document}